# Small Object Detection Based on Modified FSSD and Model Compression

Qingcai Wang
*Department of Electronic Information Engineering, School of Information Engineering Nanchang University*
Nanchang, China
wangqingcai@email.ncu.edu.cn

Hao Zhang*
*College of Electronic and Information Engineering Nanjing University of Aeronautics and Astronautics*
Nanjing, China
haozhangcn@nuaa.edu.cn

Xianggong Hong*
*Department of Electronic Information Engineering, School of Information Engineering Nanchang University*
Nanchang, China
hongxianggong@ncu.edu.cn

Qinqin Zhou*
*College of Food Science and Engineering Nangjing University of Finance and Economics*
Nanjing, China
1171990121@qq.com

*Abstract*—Small objects have relatively low resolution, the unobvious visual features which are difficult to be extracted, so the existing object detection methods cannot effectively detect small objects, and the detection speed and stability are poor. Thus, this paper proposes a small object detection algorithm based on FSSD, meanwhile, in order to reduce the computational cost and storage space, pruning is carried out to achieve model compression. Firstly, the semantic information contained in the features of different layers can be used to detect different scale objects, and the feature fusion method is improved to obtain more information beneficial to small objects; secondly, batch normalization layer is introduced to accelerate the training of neural network and make the model sparse; finally, the model is pruned by scaling factor to get the corresponding compressed model. The experimental results show that the average accuracy (mAP) of the algorithm can reach 80.4% on PASCAL VOC and the speed is 59.5 FPS on GTX1080ti. After pruning, the compressed model can reach 79.9% mAP, and 79.5 FPS in detection speed. On MS COCO, the best detection accuracy (APs) is 12.1%, and the overall detection accuracy is 49.8% AP when *IoU* is 0.5. The algorithm can not only improve the detection accuracy of small objects, but also greatly improves the detection speed, which reaches a balance between speed and accuracy.

## I. INTRODUCTION

Object detection is one of the important research directions in the field of computer vision. The purpose is to identify and locate specific objects in images or videos. It is widely used in industrial fields such as autonomous driving, video surveillance, robot vision, and new retail [1]. Traditional object detection algorithms often use sliding windows to traverse the entire image and combine artificially designed image features to achieve effective detection. Authors in [2] used a fixed-size window to detect the sliding of the entire image. Literature [3] adopted the Histogram of Oriented Gradients(HOG), which used the local gradient information of the image to effectively extract the local texture features of the image, and achieved efficient pedestrian detection. [4] utilized Deformable Part-based Model (DPM), which is an extension of the HOG method and further improves the detection performance. In these traditional algorithms, the region selection strategy based on the sliding window is not specific to the image and needs to traverse the entire image, resulting in high time complexity, window redundancy, and a lot of computing power. Artificially designed features are not very robust to the diversity of objects in the image.

In recent years, with the continuous improvement and optimization of convolutional neural networks, deep learning algorithms have shown excellent results in image classification, detection [1], and segmentation [5][6]. Scholars and experts in this field generally believe that object detection algorithms based on deep learning are divided into two categories [1]. One is Two-stage approaches, such as SPP-Net [7], Fast RCNN [8], Faster R-CNN [9], Mask R-CNN [10], etc. This kind of method uses Selective Search (SS) [11] or Region Proposal Network(RPN) [9] to generate region proposals, and then classifies the proposed regions and predicts the bounding boxes. The detection speed of these methods is slow, and cannot meet the real-time requirements in practical applications. The other is One-stage, which removes the region proposal step, performs feature extraction directly in the network, and then performs classification and regression. YOLO [12] improves the detection speed, but the final detection accuracy is low. SSD [13] draws on the YOLO idea and adopts a multiscale detection method, which not only ensures the real-time detection and effectively improves the accuracy. Subsequently, One-stage detection methods are mostly based on improved versions of YOLO and SSD, such as YOLOv2 [14], DSSD [15], FSSD [16], YOLOv3 [17], FFSSD [18], *etc*.

In the process of object detection, when the object is relatively small in the field of view, the pixel size in the image or video will also be small. MS COCO [19] defines an object with a pixel size less than 32 × 32 as a small object. The detection performance of existing object detection algorithms for small objects is only half of that of large objects (pixel size larger than 96×96). The resolution of the small object itself is low, and its features are difficult to extract from the limited resolution, and the algorithm is very easy to confuse it with the

This work is partly supported by Innovation Fund for Graduate of Nanchang University under Grant CX2018145.

surrounding background and noise, so as to cause false detection or missed detection. These are the reasons for the low performance of small target detection. In addition, with the in-depth study of convolutional neural networks, the number of layers of the network is increasing. Although the deep neural network improves the accuracy of object detection, it also increases the complexity and computational cost of the model.

In the convolutional neural network, the features extracted by the deeper network has a smaller size, a larger receptive field, and richer high-level semantic information, which is more conducive to the detection of large objects. The feature map extracted by the shallower network is generally larger in size, smaller in the receptive field, and has low-level detailed semantic information, which is more helpful for the detection of small objects. Inspired of the idea of FPN [20], FSSD integrates the features of Conv4_3, FC7 and Conv7_2 layers to improve the detection performance of small objects to a certain extent. However, we found that the retained semantic information of the feature map corresponding to the Conv7_2 layer after up-sampling by bilinear interpolation is not friendly to small objects. In contrast, the Conv5_3 layer retains more characteristic information. The features acquired by this layer are more conducive to the detection of small objects, and will not bring more noise [18].

In response to the above-mentioned small object detection problems and in order to make the algorithm better applied to embedded mobile devices, we proposed a small object detection algorithm based on FSSD [16] and model compression. The main ideas are as follows: 1) Feature fusion with different convolutional layers is utilized to make full use of the semantic information of different layers to obtain more small target information; 2) in the training strategy, the batch normalization is used to improve the performance of target detection training, and then L1 regularization is added to the BN layer for sparse training to better perform model pruning; 3) The net slimming algorithm is used to prune the network, and the pruned model is fine-tuned to obtain the final model. The experimental results show that the improved network greatly improves the detection ability of small objects. The model after pruning maintains accuracy while also greatly improving the detection speed.

## II. THE PROPOSED METHOD

As shown in Fig.1, we proposed a modified FSSD (MFSSD) by converting the original feature fusion layer into new feature layers for the better detection performance for small objects. Moreover, in order to further improve the detection ability of small objects, we increase the proportion of the feature number of features for small objects after fusion. To accelerate to training of the proposed model, BN and large batch size are utilized. Using BN and large batch size in the object detection model can significantly improve the detection performance of the model without affecting the model inference speed.

### A. Feature Fusion

Instead of using the multi-scale features independently in SSD [13], FSSD combined the multi-scale features from Conv4_3, FC7 and additional Conv7_2 to improve the detection performance. However, the integrated layers have been demonstrated that they will bring external noise into the fused features [18]. Thus, we combine the features from the original Conv4_3, Conv5_3 and FC7 layers from VGG-16 [21] to get better and robust features for small object detection. The feature fusion process can be expressed as

$$X_f = C\{X_{Conv_{4\_3}}, X_{Conv_{5\_3}}, X_{FC7}\} \qquad (1)$$

where $X_f$ represents the fused features, and $X_\_$ are the features from Conv4_3, Conv5_3 and FC7 layers, respectively. Moreover, the features from the shallow layer Conv4_3 are response for the small objects. Thus, we increase the channel number of shallow layer, and we set the channel number of Conv4_3 layer to 512, and the ratio among the fused features becomes 2 : 1 : 1.

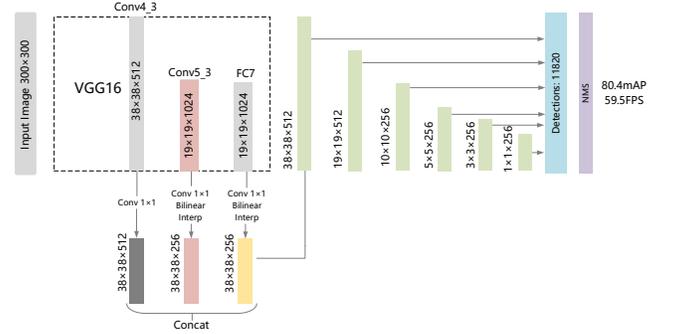

Fig. 1. The proposed MFSSD model for small object detection.

### B. Batch Normalization

Batch Normalization [22] (BN) is a method used to accelerate neural network training. Neural network training usually adopts gradient descent method. Although its training is simple and efficient, it requires manual selection of parameters, such as parameter initialization, learning rate, weight attenuation coefficient, dropout ratio, etc., and the selection of these parameters will directly affect the training results. Using the BN layer during training can achieve: 1) Use a larger learning rate to speed up the training of the network and make the network converge quickly; 2) Leave out Dropout or L2 regularization; 3) Remove the local response normalization layer; 4 ) Shuffle the training data to prevent samples from being repeatedly selected in a batch.

### C. Model Compression

In this paper, an effective structural pruning method [22] is adopted, which is a regular channel pruning strategy. The BN layer has been introduced in the previous article to accelerate neural network training and improve the generalization ability of the model. In addition, the BN layer has channel scaling parameters, and each scaling factor corresponds to a fixed convolution channel. When the scaling factor approaches 0, it can be considered that the corresponding channel is of lower importance. In this way, can be used as a factor of channel pruning, and it will not bring additional overhead to the network. In the process of model training, applying regular constraints on the parameters of the BN layer can make the model adjust the parameters in the direction of structural sparseness, and the value of a large number of channel scaling factors will continue to decrease. The role of the coefficient of the BN layer is similar to the switching coefficient of the information flow channel, which controls the opening and closing of the information flow channel. After completing sparse training or regularization, many of the values of will tend to 0, so that a suitable compression ratio (or pruning rate) can be selected to tailor the model to generate a streamlined model with low storage rate and increase a certain speedup. This method can be used in combination with other model compression methods (quantization, low-rank decomposition) to further improve the compression ratio. Combined with other optimization acceleration methods (TensorRT, *etc*.), it can increase the inference speed. If the loss of inference accuracy is too large, knowledge distillation can also be combined to effectively restore the lost accuracy

As shown in Fig.2, After the BN layer, each channel of the feature map will correspond to a scaling factor, and the size of the scaling factor represents the importance of the channel. During training, the network weights and scaling factors are jointly trained, and finally the channels with smaller scaling factors are removed to achieve network pruning. As shown in the following formula: $W$ is the network weight, $\gamma$ is the set scaling factor, and $\lambda$ is a hyper-parameter used to balance the loss caused by the network weight and the loss introduced by the scaling factor:

$$L = \sum_{(x,y)} l(f(x,W), y) + \sum_{\gamma \in f} g(\gamma) \quad (2)$$

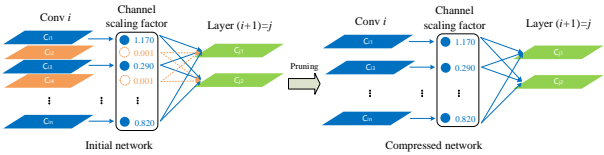

Fig. 2. Net Slimming.

The general process of model pruning is shown in Fig.3. First initialize the network, and then train the network using the channel sparse normalization method. After the network training is completed, cut out the channel with the smaller scaling factor, and then fine-tune the network. The fine-tuned network is used as a compressed network if it meets the requirements. If it does not meet the requirements, you can continue the above process.

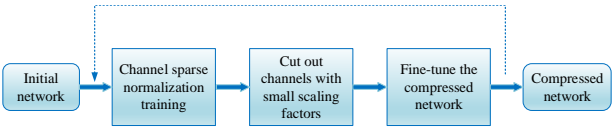

Fig. 3. The procedure of model compression through net slimming.

## III. EXPERIMENTS

### A. Datasets and Parameter Settings

The experimental data in this paper uses PSACAL VOC 2007 [23] and MS COCO [19] data sets: VOC 2007 training set data 07+12 *train-val* (16551 images), test set data *07test* (4952 images); MS COCO training set data *trainval35k* (118,278 images), test set data *test-dev* (20,288 images). All experiments of the algorithm in the article are based on the deep learning framework PyTorch [24], which is deployed on the Ubuntu 16.04 operating system. The PASCAL VOC training process uses two GPU models of GTX1080Ti (11GB video memory), CPU model is Intel® Xeon® CPU E5-2618Lv3, memory is 44GB; MS COCO training process uses 2 GPU of Titan V (12GB memory), CPU The model is Intel® Xeon® CPU E5-2628 v3, and the memory is 64GB. The GPU used in the test is GTX 1080Ti, the CPU model is Intel® Xeon® CPU E3-1230 V2 @3.30GHz, and the memory is 33GB.

The model is trained with an initial learning rate of 0.4 by the SGD optimizer with momentum and weight decay of 0.9 and 0.0005, respectively, and the learning rate is multiplied by 0.1 in the 150*th*, 200*th*, and 250*th* epochs. Moreover, the warmup strategy is adopted for the first 5 epochs [25]. Conv4_3 layer adopts 6 kinds of size preset frames to improve the detection ability of small objects.

### B. Experiment results

The results of this algorithm and algorithm in the VOC2007 test set are shown in Table II. In order to make the results more convincing, we trained SSD [13] and FSSD [16] in the same environment, with 11; 620 preset frames using the same settings as in this article. The results showed that the average accuracy of SSD and FSSD reached 77.7% and 78.6% respectively, while the detection performance of FSSDv2 reached 80.4%. After adding the BN layer, MFSSD-BN reached 81.2%, and the detection accuracy was further improved. After pruning, when the parameter amount is reduced by 33.3%, MFSSD-Prune still achieves a detection effect of 79.9%, which is higher than the detection accuracy of the original algorithm. In addition, for small objects, such as bottle and plant, MFSSD and its pruned model have shown good detection results.

In order to further verify the effect of the MFSSD proposed in this paper for small object detection, we retrained the model on MS COCO, and obtained the results as shown in Table IV. The average accuracy of MFSSD in this paper is 29.4%, which is lower than the current best one-step detection method RFB-Net (30.3%) [25], but the performance on small objects reaches the best accuracy of 12.1%. In addition, it surpasses RFB-Net (49.8% vs. 49.3%) under IoU=0.5, but the effect is not as good as RFB-Net under IoU=0.75, which shows that the model proposed in this paper still has shortcomings in the positioning of the target detection frame. The RFB-Net adopting the receptive field module can select the object frame with higher accuracy.

TABLE I. PARAMETERS COMPARISION BEFORE/AFTER PRUNING

| Model | Parameters | Size (MB) | After pruning |
|---|---|---|---|
| FSSD[15] | 34,132,960 | 130.21 | - |
| MFSSD | 33,709,664 | 128.59 | - |
| MFSSD-BN | 33,713,888 | 128.61 | - |
| **MFSSD-Pruned** | **22,785,625** | **86.92** | **66.7%** |

TABLE II.  ACCURACY COMPARISON ON PASCAL VOC2007

| model | bone | mAP | aero | bike | bird | boat | bott | bus | car | cat | chai | cow | tab | dog | horse | mbike | person | plant | sheep | sofa | train | tv |
|---|---|---|---|---|---|---|---|---|---|---|---|---|---|---|---|---|---|---|---|---|---|---|
| SSD | VGG | 77.7 | 79.9 | 85.1 | 76 | 71.5 | 54.4 | 85.8 | 85.8 | 87.6 | 58.5 | 83.5 | 76.6 | 85.1 | 87.2 | 86 | 79 | 51 | 78.8 | 77.9 | 87.6 | 77.3 |
| SSD | ResNet | 77.1 | 76.3 | 84.6 | 79.3 | 64.6 | 47.2 | 85.4 | 84 | 88.8 | 60.1 | 82.6 | 76.9 | 86.7 | 87.2 | 85.4 | 79.1 | 50.8 | 77.2 | **82.6** | 87.3 | 76.6 |
| DSSD | ResNet | 78.6 | 81.9 | 84.9 | 80.5 | 68.4 | 53.9 | 85.6 | 86.2 | **88.9** | 61.1 | 83.5 | **78.7** | 86.7 | 88.7 | 86.7 | 79.7 | 51 | 78 | 80.9 | 87.2 | 79.4 |
| MDSSD | VGG | 78.6 | 86.5 | 87.6 | 78.9 | 70.6 | 55 | 869 | 870 | 88.1 | 58.5 | 84.8 | 73.4 | 84.8 | 89.2 | 88.1 | 78 | 52.3 | 78.6 | 74.5 | 86.8 | 80.7 |
| FSSD | VGG | 78.6 | 81.4 | 86.3 | 77.1 | 71.7 | 57.2 | 85.8 | 87.1 | 88.4 | 61.5 | 85.6 | 75.9 | 86.8 | 87.5 | 87.3 | 79.5 | 50.5 | 75 | 78.5 | 88.3 | 79.8 |
| MFSSD | VGG | 80.4 | 83.1 | 88 | 79.8 | 74.9 | 60.1 | 87.6 | 87.7 | 88.1 | 66 | 85.4 | 77.5 | 87 | 88.3 | 88 | 82.1 | 57.7 | 79.2 | 78 | 88 | 80.7 |
| MFSSD-B | VGG | 81.2 | 85.5 | 87.9 | 80 | 74.6 | 63.7 | 88.3 | 87.9 | 88.8 | 67.3 | 87.6 | 76.4 | 87.3 | 89.1 | 86.8 | 82.3 | 58.8 | 82.9 | 79.9 | 88.3 | 80.8 |
| MFSSD-P | VGG | 79.9 | 82 | 85.8 | 78.4 | 73.8 | 61.5 | 85.8 | 87.8 | 88.3 | 65.9 | 85.8 | 77 | 85.9 | 88 | 87.8 | 81.6 | 55.6 | 78.9 | 80.7 | 85.7 | 80.7 |

TABLE III.  ACCURACY COMPARISON ON MS COCO

| model | bone | Avg. Precision, IoU | | | Avg. Precision, Area: | | | Avg. Recall, #Dets: | | | Avg. Recall, Area: | | |
|---|---|---|---|---|---|---|---|---|---|---|---|---|---|
| | | 0.5:0.95 | 0.5 | 0.75 | S | M | L | 1 | 10 | 100 | S | M | L |
| FSSD | VGG | 27.1 | 47.7 | 27.8 | 8.7 | 29.2 | 42.2 | 24.6 | 37.4 | 40 | 15.9 | 44.2 | 58.6 |
| MDSSD | VGG | 26.8 | 45.9 | 27.7 | 10.8 | 27.5 | 40.8 | 24.3 | 36.6 | 38.8 | 15.8 | 42.3 | 56.3 |
| DSSD | ResNet | 28 | 46.1 | 29.2 | 7.4 | 28.1 | **47.6** | 25.5 | 37.1 | 39.4 | 12.7 | 42 | 62.6 |
| RefineDet | VGG | 29.4 | 49.2 | 31.3 | 10 | 32 | 44 | - | - | - | - | - | - |
| RFB-Net | VGG | **30.3** | 49.3 | **31.8** | 11.8 | 31.9 | 45.9 | - | - | - | - | - | - |
| MFSD | VGG | 29.4 | **49.8** | 30.6 | **12.1** | **33.5** | 44.1 | **26** | 39.3 | 41.9 | **18.7** | 48.1 | 59.7 |

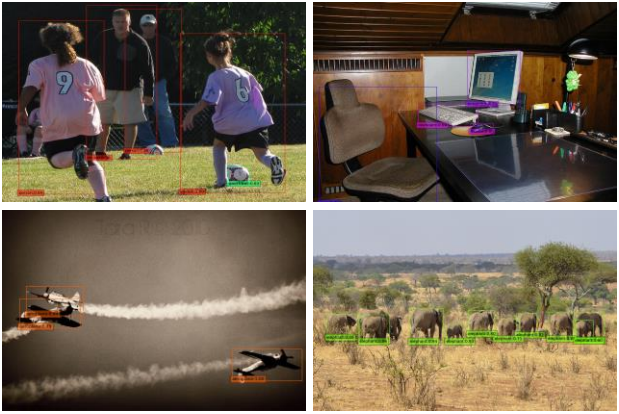

Fig. 4.  The detection result of the proposed MFSSD.

TABLE IV.  COMPARISION OF SPPED N PASCAL VOC2007

| Model | bone | mAP | FPS | #Proposals |
|---|---|---|---|---|
| SSD | VGG | 77.4 | 72 | 8732 |
| MDSSD | VGG | 78.6 | 38.5 | 8732 |
| FSSD | VGG | 78.8 | 65.8 | 8732 |
| SSD | VGG | 77.7 | 70.8 | 11620 |
| FFSSD | VGG | 78.6 | 60.8 | 11620 |
| RFB-Net | VGG | **80.5** | 49 | 11620 |
| MFFSD | VGG | **80.4** | 59.5 | 11620 |

| Model | bone | mAP | FPS | #Proposals |
|---|---|---|---|---|
| MFSSD-P | VGG | 79.9 | **79.5** | 11620 |

In order to verify the efficiency of the algorithm proposed in this article, it is compared with SSD, FSSD, and RFB-Net in the same hardware environment. The image input resolution is 300×300, and the GPU model is GTX 1080Ti. The results are shown in Table III. The running speed of the pruning model MFSSD-Prune can reach 79.5FPS. Although the accuracy is slightly lower than that of RFB-Net, it has an absolute advantage in speed. In the end, accuracy and speed are well balanced. In order to more intuitively show the effect of the algorithm proposed in this article on small object detection, Fig.4 visualizes some of the test results on MS COCO test-dev.

IV.  CONCLUSION

Small objects have relatively low resolution, the unobvious visual features which are difficult to extract, so the existing object detection methods can't effectively detect small objects, and the detection speed and stability are poor. So, this paper proposes a small object detection algorithm MFSSD based on FSSD, meanwhile, in order to reduce the computational cost and storage space, pruning is carried out to achieve model compression, namely FSSDv2-Prune. Firstly, the semantic information contained in the features of different layers can be used to detect different scale objects, and the feature fusion method is improved to obtain more information beneficial to small objects; secondly, batch normalization layer is introduced to accelerate the training of neural network and make the model sparse; finally, the model is pruned by scaling factor to get the corresponding compression model. The experimental results show that the algorithm can not only improve the detection accuracy of small objects, but also greatly improves the detection speed, which reaches a balance between speed and accuracy.

To further improve the test speed of the model for small object detection, some other techniques, such as quantization, can be utilized with model compression together. In our future work, we will try to run our model in embedded devices, with the assistance of model compression and quantization by using platforms such as TensorRT.